\documentclass[nonacm]{acmart}
\usepackage{multirow}
\usepackage{balance}
\usepackage{amsmath}

\usepackage{xcolor} 
\newcommand{\new}[1]{\textcolor{black}{#1}}

\AtBeginDocument{%
  \providecommand\BibTeX{{%
    \normalfont B\kern-0.5em{\scshape i\kern-0.25em b}\kern-0.8em\TeX}}}

\setcopyright{acmcopyright}
\copyrightyear{2023}
\acmYear{2023}
\acmDOI{XXXXXXX.XXXXXXX}

\acmConference[TSAS]{Special Issue on Machine Learning and Location Data (under revision)}{Winter 2023}{manuscript}
\acmPrice{15.00}
\acmISBN{978-1-4503-XXXX-X/18/06}

\begin{document}

\title{Building Floorspace in China: A Dataset and Learning Pipeline}

\author{Peter Egger}
\email{pegger@ethz.ch}
\orcid{0000-0002-0546-1207}
\affiliation{%
  \institution{Chair of Applied Economics, ETH Zurich; CEPR; CESifo; Nottingham Centre for Research on Globalisation and Economic Policy (GEP)}
  \streetaddress{Leonhardstrasse 21}
  \city{Zurich}
  \country{Switzerland}
  \postcode{8092}
}

\author{Susie Xi Rao}
\email{srao@ethz.ch}
\orcid{0000-0003-2379-1506}
\affiliation{%
  \institution{Chair of Applied Economics; Institute for Computing Platforms, ETH Zurich}
    \streetaddress{Leonhardstrasse 21}
  \city{Zurich}
  \country{Switzerland}
  \postcode{8092}}

\author{Sebastiano Papini}
\email{spapini@ethz.ch}
\orcid{0000-0002-7774-1726}
\affiliation{%
  \institution{Chair of Applied Economics, ETH Zurich}
  \streetaddress{Leonhardstrasse 21}
  \city{Zurich}
  \country{Switzerland}
  \postcode{8092}
}

\renewcommand{\shortauthors}{Egger, Rao and Papini}

\begin{abstract}
This paper provides a first milestone in measuring the floorspace of buildings (that is, building footprint and height) for \new{40 major Chinese cities}. The intent is to maximize city coverage and, eventually provide longitudinal data. Doing so requires building on imagery that is of a medium-fine-grained granularity, as larger cross sections of cities and longer time series for them are only available in such format. We use a \new{multi-task object segmenter approach to learn the building footprint and height in the same framework in parallel}: (1) we determine the surface area is covered by any buildings (the square footage of occupied land); (2) we determine floorspace from multi-image representations of buildings from various angles to determine the height of buildings. We use Sentinel-1 and -2 satellite images as our main data source. The benefits of these data are their large cross-sectional and longitudinal scope plus their unrestricted accessibility. \new{We provide a detailed description of our data, algorithms, and evaluations. In addition, we analyze the quality of reference data and their role for measuring the building floorspace with minimal error. We conduct extensive quantitative and qualitative analyses with Shenzhen as a case study using our multi-task learner. Finally, we conduct correlation studies between our results (on both pixel and aggregated urban area levels) and nightlight data to gauge the merits of our approach in studying urban development. Our data and codebase are publicly accessible under \url{https://gitlab.ethz.ch/raox/urban-satellite-public-v2}.} 
\end{abstract}



\received{15 May 2023}

\maketitle

\section{Introduction}
A key challenge in the social sciences dealing with macro- and meso-regional development is \new{the complete or at least vast lack of large-cross-sectional and longitudinal coverage of high-quality data. One would wish to use such data to determine the extent, location, and, eventually, the determinants and consequences of such development.} One domain in the limelight of understanding development is the role of the agglomeration of people in space – at a macro level between cities and metropolitan areas and their hinterland, and at a meso and micro level within cities and neighborhoods. Measuring agglomeration well at the macro, meso, and micro levels in space is important for learning about its fundamental role in the course of development, for resource use and saving, for gauging the benefit of infrastructure developments in general and of various types thereof \new{in particular}, etc.

However, large sets of data covering many and, ideally all, neighborhoods on various scales and other data to compare with are not always available. \new{And if they are available, then mainly for where the development process has already matured, namely in industrialized countries such as the United States \citep{doi:10.1080/0042098984510, 10.2307/25098858} , Germany  \citep{RePEc:nbr:nberwo:20354}, or England \citep{10.1093/qje/qjaa014}.} In the light of recent interest in cities in developing countries \citep{10.2307/3986401,Tsivanidis2018TheAA,RePEc:anr:reveco:v:12:y:2020:p:273-297}, one would like to have such data available over a longer time span in countries, which are situated at the steep ascent of the economic growth and catching-up \new{trajectory}, as causation is likely easier to achieve in contexts, where big changes occur, and they do so to a sufficiently heterogeneous extent in data that had been collected in a standardized way. The latter is something that this paper aims at accomplishing by utilizing modern methods in computer vision in conjunction with satellite imagery in remote sensing from various sources on the housing stock in a large number of cities in China.

China is an economy that fits the purpose very well. To start with, it grew at a much faster rate than the data-rich industrialized countries since its opening up to foreign trade \new{in 1978, and in particular since the early 1990s.} Hence, it grew very rapidly in a time span when remote sensing data were available. Second, it has undergone a substantial structural change from an agricultural economy to one of the leading manufacturers on the globe, and more recently a big surge in certain services, as is suggested by the change in the overall and the field-specific enrollment rates at its universities. Hand in hand with structural change, the country witnessed a transformation in the distribution of its population between rural areas and agglomerations (cities and metropolitan areas). The latter is partly reflected in the reforms of the household registration system (called \textit{hukou}), which essentially grants legal access to home ownership, education, or the health system, but also beyond (e.g., in terms of illegal migrant work in agglomerations, which is to a certain extent tolerated by the authorities as long as it benefits the economy). Hence, over the past four decades a large surge of the demand and supply of floorspace in Chinese cities emerged.

When speaking of agglomeration, demographers and other social scientists typically think of the density of people in space. In most economies and jurisdictions, detailed data on the residents in the meso or micro space are surprisingly scarce, and they can often only be garnered from sources where households somehow disclose their presence (e.g., official work records, official purchases of homes, etc.). Yet, even in the United States, an illegal migrant could work on the black market and rent a home without going on record regarding their residence, and in some agglomerations the mass of off-record individuals might be non-trivially large. Moreover, one might speak of agglomerations from the viewpoint of production, and then the space occupied by it would be of interest. In that case, official data might also be problematic for a host of reasons. For example, register data of companies typically list the parent company and its address. This is exactly problematic for those companies that are said to account for the lion’s share in production, namely multi-plant and –site units. Also, such records may list the core activity. However, with large manufacturing companies, for example, the headquarters are often in city centers, while the production operations are in the outskirts or the hinterland of cities. The latter suggests that production-related agglomeration aspects are likely mismeasured, as are residence-related agglomeration aspects of households. Altogether, this creates a demand for an alternative route of measurement for the agglomeration phenomenon.

\new{This paper provides the first milestone in this context by delivering an open-source rich dataset on building floorspace (that is, building footprint and height). The \new{current} data set is composed of 40 cities where substantial infrastructure changes occurred between 2014 and 2017.} \new{The intent is to broaden the scope both in terms of cross-sectional and longitudinal coverage. We} build a pipeline to collectively learn building footprint and height in \new{a multi-task learner}.  The challenges of this attempt are the following.
\begin{enumerate}
    \item \textbf{Data sources and collection.} \new{Satellite imagery at a high resolution together with trustworthy reference data on the content of that imagery (e.g., buildings, heights of buildings, etc.) only exist for selected cities if not even only subregions within them. Hence, such imagery as well as reference data often require researchers to perform substantial pre-processing steps in both the satellite imagery as well as the reference data. The quality of such pre-processing certainly impacts the detection results and model validity.}
    \item \textbf{\new{A joint learning approach for} building footprint and height.} Deep learning methods have fundamentally transformed the ways people use remote sensing data. \new{New techniques have emerged on how to combine various objectives with multi-task or two-stage learning procedures. However, the merits of such joint approaches for the measurment of floorspace occupation and development for the study of urban change have yet to be assessed.} 
    \item \textbf{Evaluation of prediction quality.} How to interpret and validate the model predictions of urban change and the location and height of buildings is a key question in this context. \new{Evaluating predictions solely based on one metric or model can be misleading. It appears crucial to consider various data sources (e.g., on population and infrastructure density, nightlight data), metrics, and models to provide a comprehensive understanding of urban dynamics.}
\end{enumerate}

The \new{present} paper delivers a rich dataset on the expansion of China’s floorspace. We first collect \new{daylight satellite data of Sentinel at the scale of $10\times 10$ meters and corresponding three-dimensional (3D)} reference data in 40 cities. We then use a multi-task learning approach to gauge the building footprint and height in the same framework: (1) we determine the surface area covered by any buildings (the square footage of occupied land); (2) we determine the height of buildings from their imagery. \new{Finally, we evaluate various U-Net architectures popular in computer vision to predict these two outcomes. We empirically investigate the prediction quality of floorspaces in Shenzhen as a case study, where we inter alia assess the correlation of floorspace density with nightlight luminosity data. Our data and codebase are publicly accessible at} \url{https://gitlab.ethz.ch/raox/urban-satellite-public-v2}.

\new{The key contributions can be summarized as follows. We provide a detailed pipeline to study the building floorspaces in 40 Chinese cities. We indicate in a \new{robustness check using nightlight data how the obtained floorspace density predictions} can be used to analyze microregional agglomeration patterns within cities. E.g., the latter helps speaking to the question of mono- versus polycentric structures of cities, and it potentially helps identifying the changing location of city cores over time. The present work serves as a foundation for our future efforts to determine the evolution of floorspace over time for all the Chinese cities over a longer time span.}


\section{Building footprint recognition and building height estimation}

\subsection{Data preparation} \label{sec:data}
We use the Sentinel satellite imagery as our main data source and benefit from its \new{open-access and free-of-charge} policy. The image sets include both Sentinel-1 (Synthetic-aperture radar, SAR) and Sentinel-2 (multi-spectral optical) products which both provide near global coverage and high spatial resolution. We implement the \new{data download and image processing through Google Earth Engine API for a total of 40 cities in China, where substantial infrastructure changes took place between 2017 and 2022.}\footnote{The 40 cities in our sample are Baoding, Cangzhou, Changsha, Chengdu, Foshan, Fuzhou, Guiyang, Hongkong, Hohhot, Huizhou, Jiaxing, Langfang, Linyi, Nanchang, Ningbo, Qinhuangdao, Quanzhou, Sanya, Shaoxing, Shengyang, Shenzhen, Suzhou, Taizhou (Zhejiang), Taiyuan, Tangshan, Wenzhou, Weifang, Weihai, Wuhu, Wuxi, Xiamen, Xi'an, Xuzhou, Yantai, Yangzhou, Yinchuan, Zaozhuang, Zhongshan, Zhuhai, and Zhenjiang.} These cities are selected from the reference data  \new{that are crowd-sourced and widely used in academic remote-sensing communities, see \citet{CAO2021112590}}. 

For Sentinel-2 optical images, we select the level 2A product which provides orthoimage Bottom-Of-Atmosphere (BOA) corrected surface reflectance data. The bands we use contain B2 (blue), B3 (green), B4 (red), and B8 (NIR), with a spatial resolution of $10\times 10$ meters. To reduce the impacts of cloud occlusion, we first filter out images with more than 60\% cloud coverage and further apply the cloud mask by making use of the additional Sentinel-2 cloud probability image collections.\footnote{See \url{https://developers.google.com/earth-engine/tutorials/community/sentinel-2-s2cloudless} (last accessed: March 3, 2023).} We query the data based on the above condition from 2017 to 2021 and take the mean values from multiple returned images to get a single image for each year, covering the area of interest for each city.\footnote{Note that the results we report in Section~\ref{sec:eval} are obtained using satellite images from 2017, as it is conjectured that the crowd-sourced reference data is from 2017. We will use 2017-2021 in future work in a similar setting, especially when we have better quality ground-truth data at hand, as discussed in Section~\ref{sec:future}. Note we exclude data in 2022 at this stage because some significant changes to the calibration (band-dependent offset) were made in 2022 \url{https://developers.google.com/earth-engine/datasets/catalog/sentinel-2} (last accessed: April 23, 2023).}

For Sentinel-1 radar images, we select VV and VH polarizations acquired from a C-band SAR sensor with combined ascending and descending orbit directions to collect maximum information content. Data are queried over the years from 2015 to 2021, and for each year, the same image collections reduction technique as for Sentinel-2 is applied.


\subsection{Method: multi-task learning}

\new{We promote a multi-task learning approach that uses neural architecture to segment building footprints and predict building height. We adopt U-Net \cite{ronneberger2015u} with various backbones. The vanilla U-Net architecture described in \cite{ronneberger2015u}  does not rely on any specific pre-trained backbone networks. Instead, it utilizes its own encoder-decoder convolutional structure with skip connections. It was originally developed for biomedical image segmentation. It became later popular in urban studies for land cover classification or for the extraction of roads or buildings from satellite imagery (c.f.~\citet{unetcomparing}). We test various U-Net backbone variants, including EfficientNet-B4 \citep{tan2019efficientnet}, ResNet-34 \citep{he2016deep}, and SE-ResNet50 \citep{hu2018squeeze}. We modify these architectures to take multiple inputs and produce multiple outputs. } 

\new{We have manually evaluated the data quality in the crowd-sourced dataset. There are substantial differences in data quality both between and within cities. We focus on the 40 cities with relatively good reference data quality. This equates to 1035 mosaic tiles that we have downloaded from Google Earth Engine (Sentinel 1- and -2) using our script on GitLab. To prepare training data, we partition these images into $256\times 256$ grids of respective tiles. The reference data is then rasterized form polygons per building to the same $10\times 10$ meter grid resolution that Sentinel imagery comes in. We filter out tiles that have less than 10\% of its pixels belonging to buildings. The train/validation split ratio is 90\%/10\%. For the test set, we use all the images of the city of Shenzhen. The reasons using the latter city are the following. (1) Shenzhen has witnessed large infrastructural and urban changes in the past three decades and the city provides many open-source data such as firm registry and 3D building points through the Shenzhen Municipal Government Data Open Platform \citep{sz_gov}. (2) Compared to other cities, Shenzhen has been well studied in previous work modelling the 3D building floorspace from 1986 to 2017 \citep{yuetal2021}. The latter can be used to compare our results with, and it provides an excellent base for model validation. This is important, as we plan to extend our pipeline to other major Chinese cities where reference data are not available. Hence, we need a pipeline that is robust when tested in out-of-sample cities. } 

\new{We learn the building footprint and height in a multi-task model. For footprint, we predict if the area of interest is covered by any buildings or not. For building height, we predict and learn the height conditional the area being covered by buildings in a regression-based model. For more stable training, instead of directly predicting absolute building height, we predict normalized height. For normalization, we divide the height values by 400 meters, because the maximum height of any building in our data set is below 400 meters. We use cross-entropy for building-footprint prediction and smooth L1 loss for height prediction. To calculate the loss for height prediction, we mask out the pixels that do not belong to buildings. The final loss is a linear combination of the two losses with a weight of 0.1 for building footprint prediction loss and a weight of 1 for building height regression loss. Note that we have experimented with data-augmentation techniques discussed in Section~\ref{sec:data-aug} and various task coefficients for area classification and height regression. For footprint prediction, we report precision, recall, and the Dice coefficient (the harmonic mean of precision and recall); for height, we report the Mean Absolute Error (MAE), and the Root Mean Squared Error (RMSE).} 

Important hyperparameters of the respective neural architectures are multi-step learning rate that starts with 0.001 and decays by factor of 0.1 in the 50th epoch. We benchmark our algorithm in a machine with 64 units of AMD EPYC 7313 16-Core Processor, 504 GB memory, and a single unit of GPU (NVIDIA GeForce RTX 3090, 24GB memory), and the training time is about \new{15 seconds per epoch on average. We train all the architectures for 100 epochs.}

\section{Results and Evaluation}\label{sec:eval}

\new{In this section, we first evaluate the performance of U-Net with various backbones on the validation set for footprint classification and height prediction. Then, we reflect on our choice of model and share results of ablation studies in a two-stage model, log normalization and data augmentation. Finally, we provide 50 case studies of an estimated floorspace based on the proposed model vs.~the reference data for the city of Shenzhen.} 

\subsection{Quantitative analysis}
\begin{figure}[!t]
\includegraphics[width=1\linewidth]{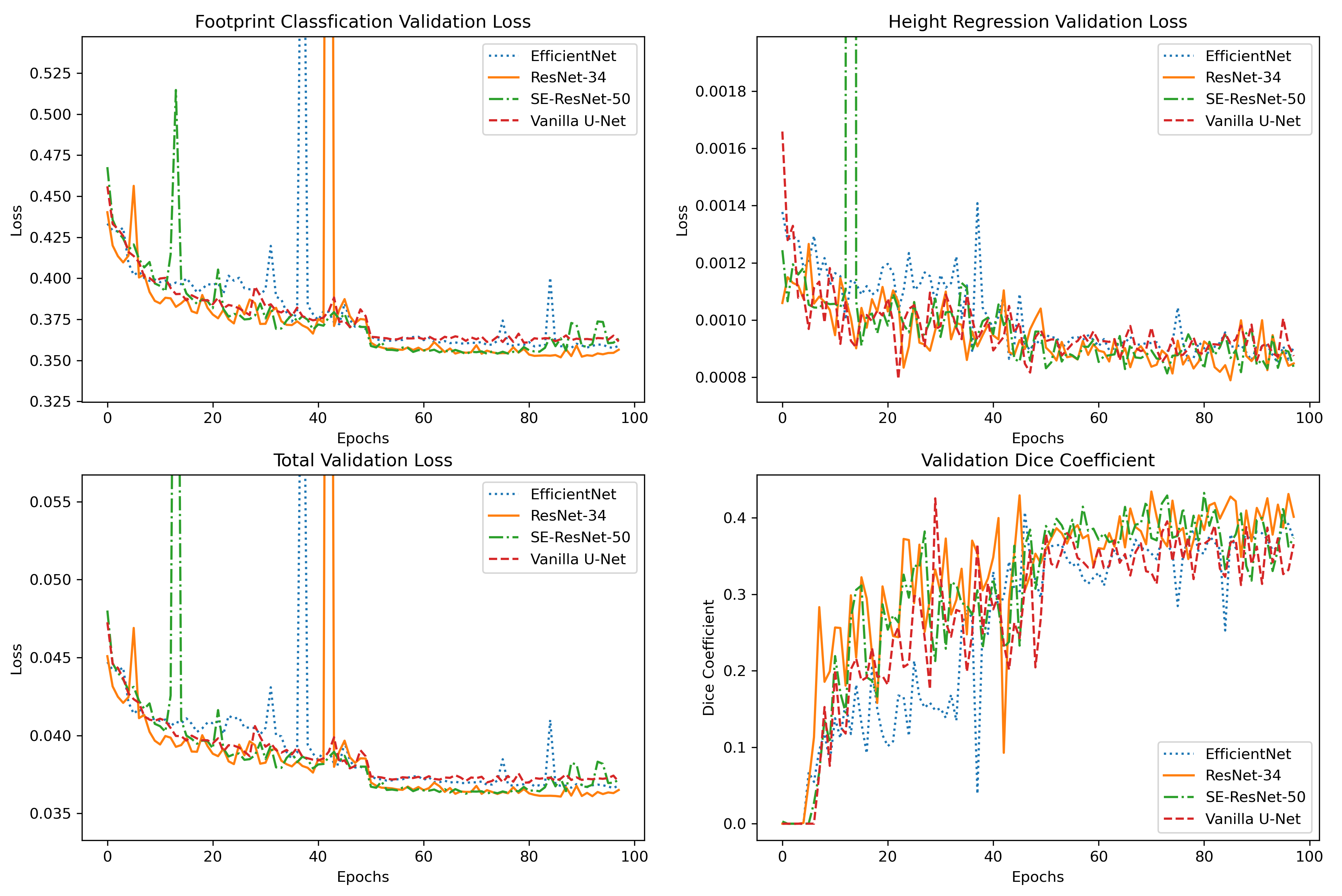}
\caption{Results of Various U-Net Backbones in the Validation Set.}
\label{fig:unet-variants} 
\end{figure}

\textbf{Backbone selection.}
\new{We use the overlay of Sentinel-1 and -2 as the input of our multi-task learner (see Appendix \ref{app:footprint} for our ablation studies on data sources). We first investigate various U-Net backbones in our multi-task learning setting. In Figure \ref{fig:unet-variants} we illustrate the performance of these various U-Net backbones on the validation set (10\% of the data) in terms of the Dice coefficient and the respective losses. From this comparison, we select ResNet-34 as our U-Net backbone. We illustrate its model architecture in Figure \ref{fig:model_arch} of Appendix \ref{app:resnet-34}.} 

\textbf{Evaluation metrics of building footprint and height. }
For building footprints, we quantify the performance on the pixel level. The evaluation metrics that we use are precision, recall, and the Dice coefficient. Precision and recall are calculated only on the true positives (buildings), the Dice coefficient (also known as F1) is the harmonic mean of precision and recall. \new{For learning building height, we compute the mean absolute error (MA), the mean relative error (MRE), and the root mean-squared error (RMSE). MAE is the average absolute difference between the predicted and actual heights, while MRE is the average ratio of the absolute difference to the actual height. RMSE measures the average deviation between the predicted values and the actual values in a dataset. MAE and MRE are absolute measures of model performance, indicating the average magnitude of the errors. RMSE puts more weight on larger errors and is more sensitive to outliers (e.g., higher buildings), while MAE treats all errors equally and is less influenced by outliers. } 

\begin{figure}[!t]
    \centering
    \includegraphics[width=0.7\linewidth]{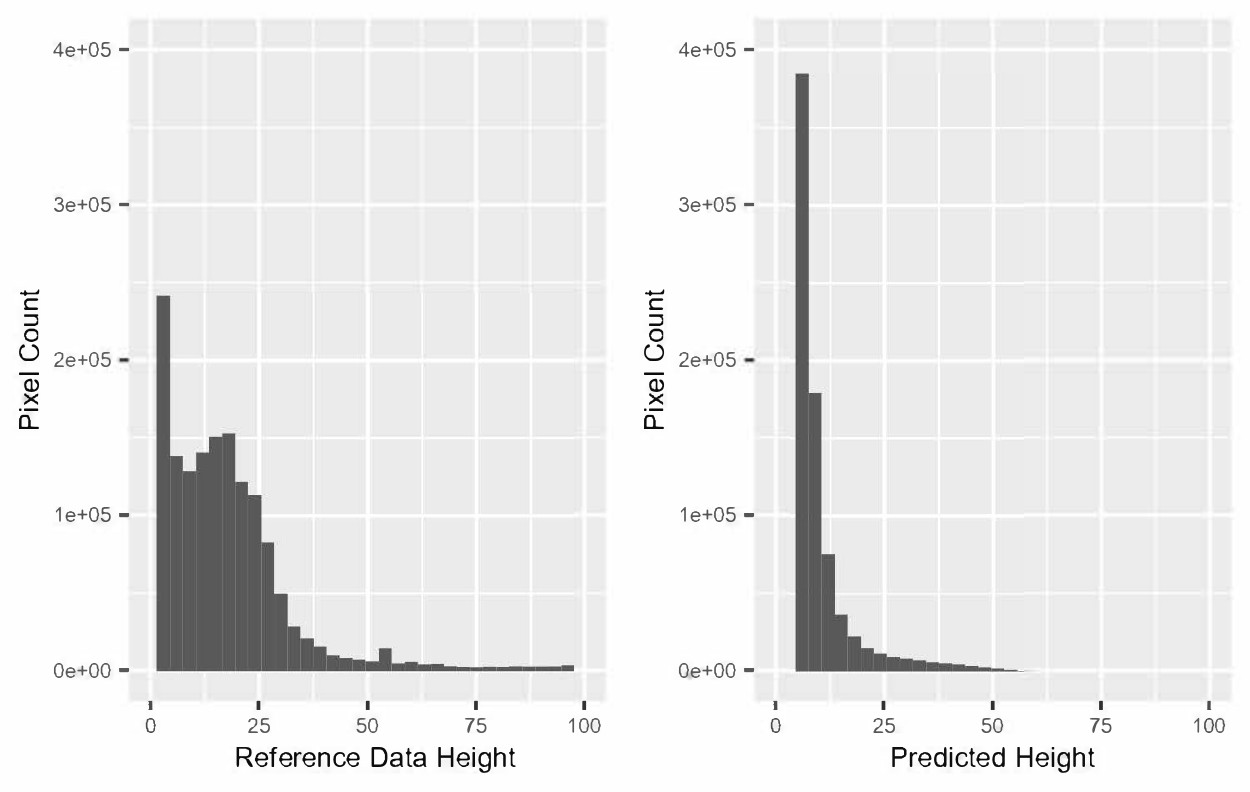}
    \caption{Histogram of the Pixel-wise Height Distribution for Shenzhen in the Trustworthy Reference Data Vs.~the Prediction Up to 100 Meters.}
    \label{fig:hist_gt_pd}
\end{figure}

\textbf{Building heights in Shenzhen.}
\new{In \citet{yuetal2021}, the authors report the reference data of 3D building floorspace in 2017 in Shenzhen. We analyze our results in Shenzhen (test data) using the classification schemes they recommend. They have high-quality  (non-public) reference data of 567,000 buildings provided by the Bureau of Planning and Natural Resources in Shenzhen. There are four types according to this classification: low-rise buildings (1–3 storeys), multi-storey buildings (4–6 storeys), mid-rise buildings (7–9 storeys) and high-rise buildings ($\geq$10 storeys).  The frequencies of the four types are 51.9\%, 31.0\%, 12.5\%, and 4.5\% in 2017. Note that this distribution is reflected in the trustworthy reference data Figure~\ref{fig:hist_gt_pd} (left panel).} 

\textbf{Trustworthy reference data in Shenzhen.}
\new{In order to have a reliable test set we select trustworthy reference data in Shenzhen, we conduct a manual comparison between our reference data and Maxar's 0.5m high-resolution satellite imagery \citep{sz_max}. We consider areas where roughly more than 90\% of the buildings were correctly marked as relatively trustworthy and used them for our evaluation.}

\textbf{Our predictions in Shenzhen.}
\new{In Table~\ref{tab:models} we present the results of the best performer, U-Net with a ResNet-34 backbone in predicting the trustworthy reference data in Shenzhen. The MAE value of 3.4566 indicates that, on average, the model's predictions have an absolute difference of approximately 3.5 meters (about 1 storey) from the actual height values. The RMSE value of 9.8385 indicates the square root of the average squared differences between the predicted and actual height values. It provides a measure of the typical magnitude of the errors made by the model. In this case, it suggests that, on average, the model's predictions have an error of approximately 9.8385 meters. The RMSE is sensitive to larger errors (here, higher buildings) and penalizes them more than the MAE. The MAE and RMSE values indicate some level of error in the predictions, with RMSE being higher due to its sensitivity to outliers.}

\new{We classify the buildings following \citet{yuetal2021} and calculate the MRE for each class. For low-rise buildings (1–3 storeys) the MRE is 1.3159, for multi-storey buildings (4–6 storeys) the MRE is 0.9405, for middle-rise buildings (7–9 storeys) the MRE is 0.8842, and for high-rise buildings ($\geq$10 storeys) the MRE is 0.8511. The MRE values provided for each class of buildings indicate the average relative difference between the predicted heights and the actual heights. These numbers suggest that the model tends to \textit{overestimate} the heights of low-rise buildings, with the predicted heights being, on average, around 31.59\% higher than the actual heights. The model's predictions for multi-storey buildings, middle-rise and high-rise buildings are more accurate, with the predicted heights being, on average, around 6.05\%, 11.58\%, 14.89\% different from the actual heights, respectively.}

\begin{table*}[!t]
\begin{tabular}{llllll}
\toprule

      & \multicolumn{3}{c}{\textbf{Footprint}} & \multicolumn{2}{c}{\textbf{Height}}       \\
      \cmidrule(lr){2-4} \cmidrule(lr){5-6}
      & \textbf{Precision} & \textbf{Recall} & \textbf{Dice (F1)} & \textbf{MAE} & \textbf{RMSE} \\ 
      \cmidrule(lr){2-4} \cmidrule(lr){5-6}
 
ResNet-34 &  0.5881 & 0.2959  & 0.3872  & 3.4566 & 9.8385    \\ 
\bottomrule
\end{tabular}
\caption{Results for Multi-Task Learner U-Net with ResNet-34 Backbone on the trustworthy test set (Shenzhen City).}
\label{tab:models}    
\end{table*}

\new{To illustrate the predicted versus reference-data height distributions of buildings, we show in Figure~\ref{fig:hist_gt_pd} the histograms of the distribution of predicted and targeted heights per pixel for Shenzhen. The x-axis is the building height in meters, the y-axis is the pixel count. Both histograms are capped at 100 meters. Based on the MAE and MRE scores, we see that our model underestimates average heights, while overestimating low-rise buildings of 1-3 storeys, as is reflected in the MRE scores of the four building categories. Our predictions miss high-rise buildings beyond 60 meters. However, as we will see later in the case studies (Section \ref{sec:case}), the reference data are not perfect either. Therefore, we conduct robustness checks with other types of geospatial data, such as nightlight data -- those are frequently used in urban economics -- to further validate the predictions the proposed model makes (see Section~\ref{sec:nightlight}). } 

\subsection{Discussion of model architecture} 

\subsubsection{Two-stage model}
\new{Previous work by \citet{wen2019monitoring}, who had access to a commercial data set of multi-view images from the ZY-3 satellite, used two-stage models to model the 3D building space.} \citet{CAO2021112590}, who studied 42 Chinese cities, had access to the same sources as \citet{wen2019monitoring} and learned the exact heights of the buildings from them. However, note that such multi-view images are not publicly available for China, and they are expensive to acquire. Moreover, the longitudinal coverage of these data is very limited.

The goal of the framework considered here is to (i) be able to cover all urban areas in China, (ii) cover a time span of much more than a decade, and (iii) build on open-access and free-of-charge data sources. The evidence provided by the present paper suggests that such a goal can be pursued, in particular, as long as the focus is on meso-to-marcro-level regional aggregates of grid cells with a side length of 10-20 meters. The latter is a level of sufficient resolution for most academic studies with a demographic, urban, regional, or transport focus. Apart from the difference in the targeted resolution, another key difference between the approaches in \citet{wen2019monitoring} and \citet{CAO2021112590} on the one hand and ours on the other hand is that we combine the learning of the building footprint and building heights in a single stage model.

\textcolor{black}{In the absence of multi-view imagery, we conjecture the following with regard to our two-stage model: our implementation is comparatively more efficient than the one in the aforementioned work; and the performance of the two-stage models will not differ much from a one-stage model. As a robustness check, we conducted two-stage experiments on the best setup of Sentinel 1+2, the results of which substantiate our claims. In Appendix~\ref{app:2-stg}, we present the experimental setup and results.} 


\subsubsection{Output Normalization and data augmentation}
\label{sec:data-aug}
\new{We have attempted various approaches for output normalization in our multi-task learner. By normalizing the output, one changes the importance of the building-footprint classification task and the building-height regression task. Fine-tuning the loss function can help ensuring that both tasks receive appropriate emphasis during training. Instead of using a scaling factor of 400 meters (for maximum building height in the data), we used the log scale height in the loss function. Moreover, we experimented with various coefficients (0.05, 0.1, 0.2, 1, 10) for the relative weighting of the building-height-prediction task relative to the footprint-prediction task.}

\new{Additionally, we have conducted data-augmentation experiments inspired by techniques introduced in \citet{liu2021data}. Those authors showed in their experiments that pixel transformations give similar or slightly worse performance than the baseline. They also showed that random rotation, shear X/Y, and randomly erasing some pixels could help reducing over-fitting by artificially increasing the size of the training dataset. We considered the following experiments in this spirit. We implemented random rotation with a range of [-10, 10], random affine transformation which includes shear X/Y, translation, and rotation, and random masking some pixels. In the random masking, only Sentinel-1 data were randomly masked out, because we already could have randomly missing pixel data in Sentinel-2 due to the occlusion by clouds}. Based on these experiments, we gained the following insights.

\new{Notably, data augmentation did not improve the performance with the loss scaled by dividing 400. However, augmentation helped with the log scale loss. Using only rotation performed the best. Using the affine transformation also improved the performance, but it performed worse than just rotation. The best performer using the log scale loss was SE-ResNet-50-log-reg-0.05-rotation (giving the regression task a coefficient of 0.05 using SE-ResNet-50 as the backbone with rotation as data augmentation). However, as we show in the evaluation on the city of Shenzhen (see Appendix \ref{app:weighted_loss}), the loss with a scaling factor 400 meters performs better than the log scale plus rotation with the Sentinel data.}

\subsection{Case studies}
\label{sec:case}

\begin{figure*}[!t]
    \centering
    \begin{tabular}{c}
    \includegraphics[width=1\linewidth]
    {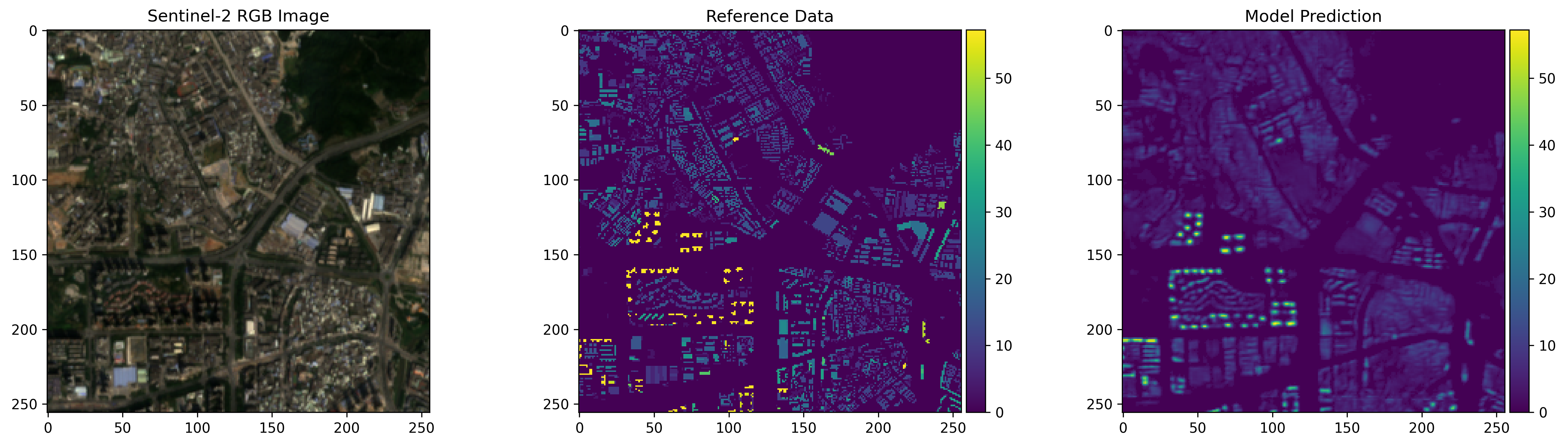} \\
    \textbf{\small (a) Overestimate of building
footprint coverage. } \\
    \includegraphics[width=1\linewidth]{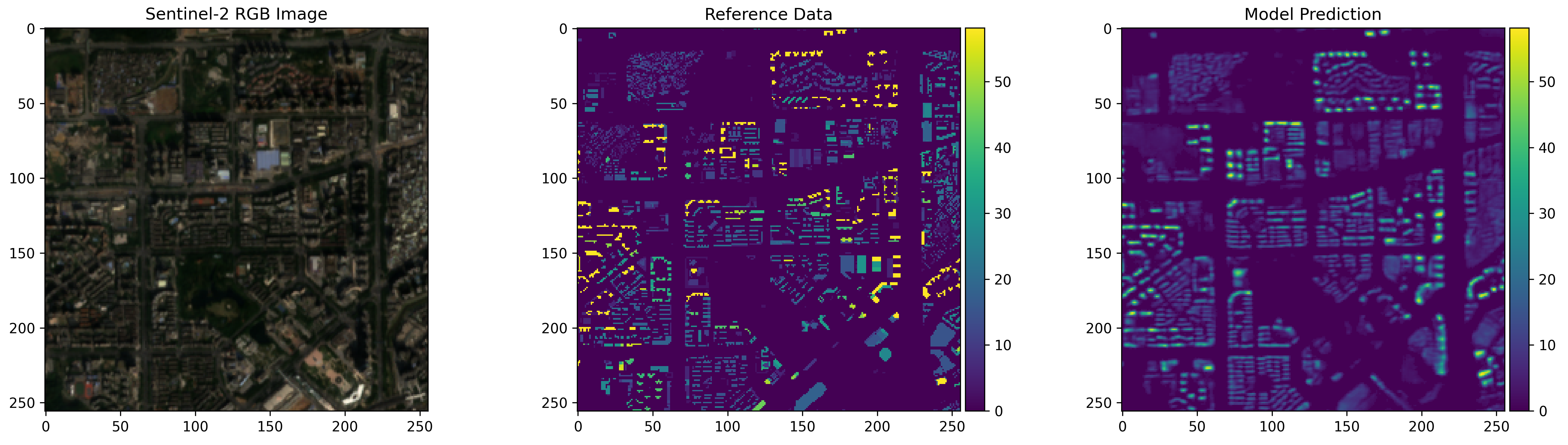} \\
    \textbf{\small (b) Underestimate of building height of high-rise range.} \\
    \end{tabular}
    \caption{Comparison of Sampled Tiles in Shenzhen. The axes represent the dimensions of the pixels, and the color represents the heights of the buildings. (a): Overestimate of building
footprint coverage, (b): Underestimate of building height of high-rise range. }
    \label{fig:case2}
\end{figure*}

\new{To further evaluate the quality of our predictions compared to crowd-sourced reference data, we conduct a manual analysis of 50 randomly selected 256x256 tiles. In one tile, each pixel represents a dimension of 10 meters obtained from our predictions within the Shenzhen area.} 

\new{We identify two primary challenges. First, in terms of footprint, all of our sampled tiles successfully identify the regions covered by the buildings; however, in 38\% of the samples, our model encounters difficulties in accurately distinguishing individual buildings that are small and located close to each other, leading to an overestimate of building footprint coverage, as shown in Figure \ref{fig:case2} (a). We believe this is largely due to the inherent limitation of Sentinel resolution in relation to the urban topology. In our designated reference area in Shenzhen, approximately 91\% of the buildings are located within a distance of less than 10 meters from their nearest neighbouring building, corresponding to the size of a single pixel. Furthermore, 54.6\% of the buildings are within 10 meters of at least two other buildings. Consequently, we propose that evaluating the prediction results at an aggregated level (discussed in Section \ref{sec:nightlight}), rather than the pixel level, would yield more meaningful insights.} 

\new{Second, from the perspective of height estimation, although 92\% of our predictions accurately portray the hierarchical structure of building heights, they consistently underestimate the height of high-rise buildings as shown in Figure \ref{fig:case2} (b). Furthermore, during our analysis, we discover that our model correctly manages to identify the buildings under construction in three of the sampled images, which have not yet been documented in the reference data. This finding indicates that the presence of such unmarked buildings could influence the evaluation process. We provide the best-performing model, the test images and the tiles of case studies in our project repository on GitLab. }

\section{Prediction Quality in Aggregated Grids and Robustness Checks with Nightlight data} \label{sec:nightlight}

Over the past two decades, night-light (NTL) satellite (remote-sensing) imagery became a workhorse measure in urban and regional studies, demography, and economics. NTL data have been used extensively to measure city structure \citep{gonzalez2018subways, lan2019spatial}, population density \citep{zhuo2009modelling, elvidge2001resolution}, economic activity and growth \citep{henderson2012measuring, baum2017roads, zhu2019assessing, gibson2021night, duan2018understanding}, and poverty and wealth \citep{elvidge2012night, asher2021development}.
The reasons for the appeal of NTL data to measure and approximate for a host of socio-demographic and -economic factors are two: (i) they are available throughout the world for a relatively long time span of several decades; and (ii) they are measured with comparable or identical precision throughout space and time. On the contrary, more exact measures of demography, economic activity, or urban and regional development are often measured at much lower frequency (e.g., census data at the decade level), in a selected way (samples are drawn with reporting thresholds; or only selected jurisdictions report specific data), and the reporting is not conducted with the same quality or at the same time across jurisdictions. A disadvantage of NTL data relative to the ones generated in this paper is that they do not come at a similarly fine-grained spatial resolution level as the Sentinel and other daylight satellite imagery. \citet{henderson2012measuring} used NTL with nearly $1000\times 1000$ grid resolution and \citet{gibson2021night}, as a more recent example, have utilised NTL on a $700\times 700$ meter resolution.

\new{Their widespread use and availability made NTL data a natural candidate to compare our floorspace predictions with. The reason is that one would expect densely populated areas (i.e., ones with high-rise buildings and a large density of floorspace per unit of covered land) to emit more light during the nighttime. In pursuit of this conjecture, we compared our floor-space predictions with the nightlight radiance in Luojia 1-01 NTL data on a $120\times 120$ meter grid for Shenzhen in 2019. Prior to investigating the correlation between building floorspaces and nightlight data, we considered the correlation between our predictions and trustworthy data in aggregated grids (see Section \ref{sec:pred-ref-grid}). This allows us to interpret our prediction quality on a scale, where urban-development indicators based on NTL data are typically computed. Then, in Section \ref{sec:nlt-check} we compare the footprint and height coverage of predictions made by our best performing model, the reference data, and nightlight data. We evaluate the merits of our floorspace estimates by our model relative to the NTL data. Specifically, the biggest merit is clearly that the predictions made available in the present paper unveil the urban density that NTL data fails to represent, as they are influenced by light emissions from other sources than buildings (e.g., traffic or illuminated infrastructure other than buildings).}

\subsection{Prediction and Trustworthy Reference Data in Aggregated Grids}
\label{sec:pred-ref-grid}

\begin{figure*}[!t]
    \centering  \includegraphics[width=0.6\linewidth]{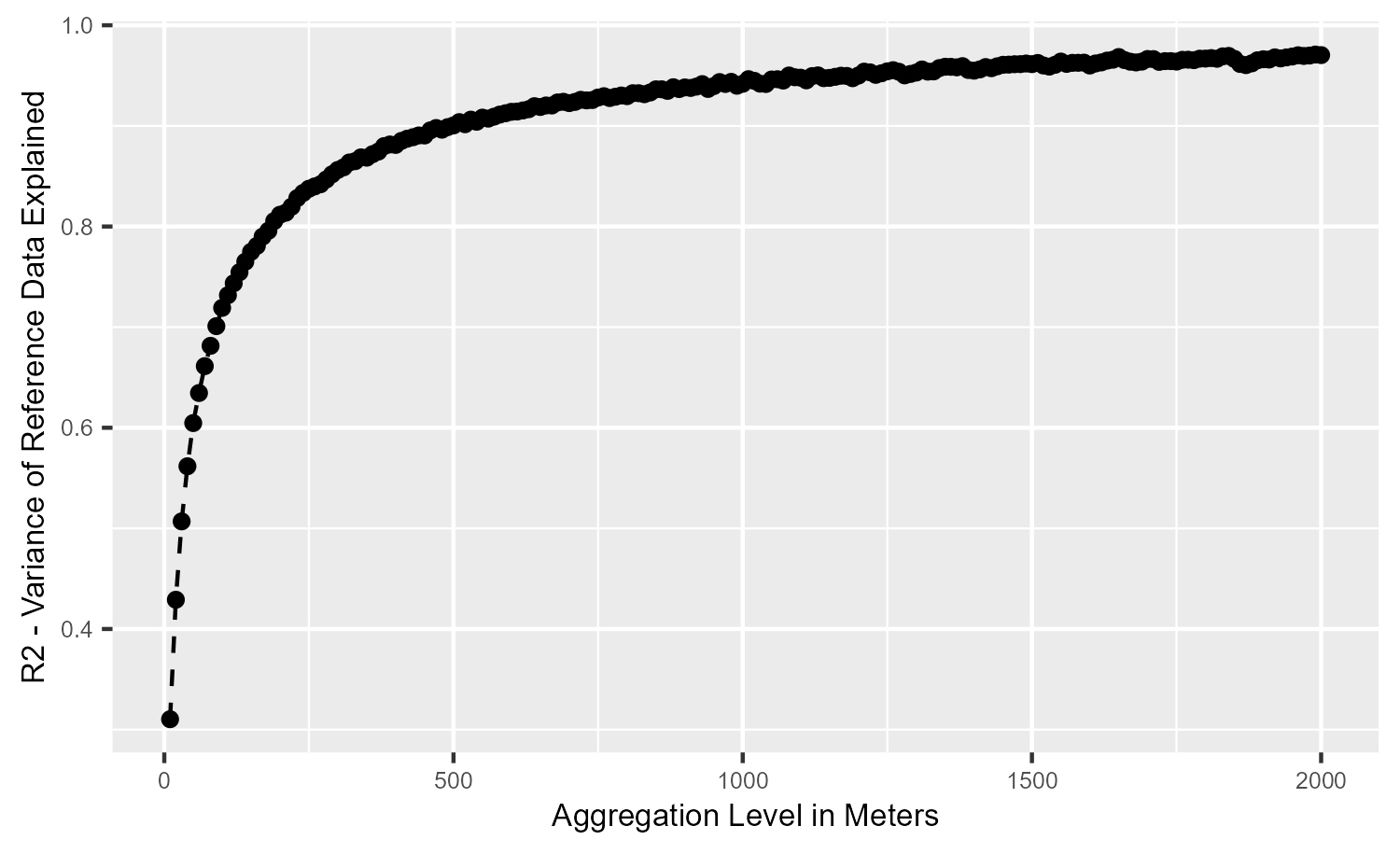}
    \caption{Share of Explained Variance of Building Height in the Reference Data by Model-predicted Building Height as a Function of the Aggregation Level. X-Axis: Side length of grid cell in meters. Y-Axis: Proportion of trustworthy reference data building height variability explained by a correlation model informed by the model prediction. }
    \label{fig:agg-corr}
\end{figure*}

\begin{figure}[!t]
    \centering
    \includegraphics[width=0.5\linewidth]{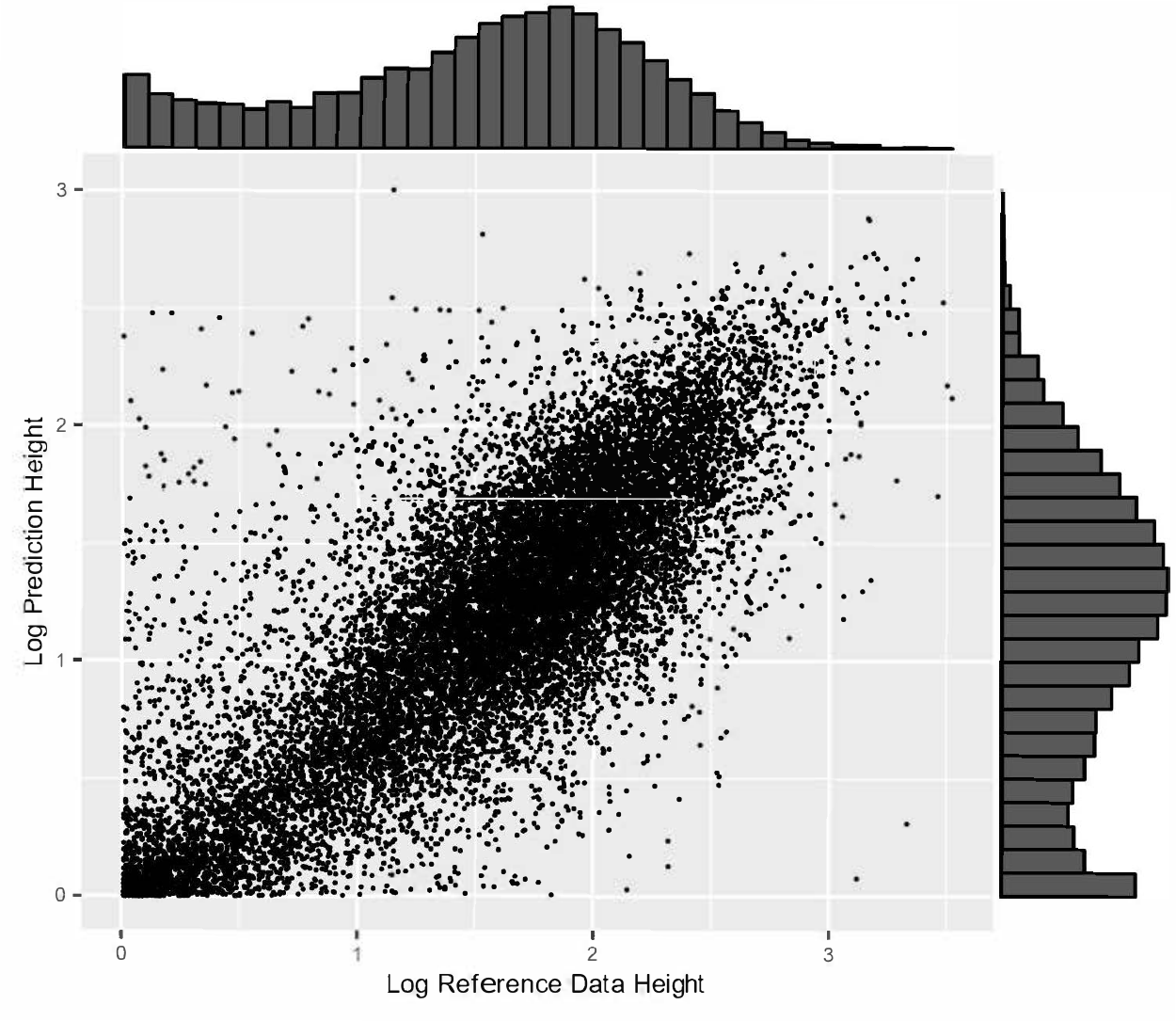}
    \caption{Scatter Plot of Aggregated to 200 Meter Cells Logarithm of Mean Height Plus One. The trustworthy reference data are shown on the x-axis and predicted height on the y-axis. Only cells with non-zero values for both dimensions are shown.}
    \label{fig:scatter_rf_pd}
\end{figure}

\new{Figure \ref{fig:agg-corr} shows the MSE of a simple correlation model between our predicted building heights and the reference data as a function of spatial aggregation (the side length of spatial grid cell) for Shenzhen in 2019. The data shown here include only the trustworthy part of the reference data for Shenzhen. The x-axis represents side lengths of the grid cells and each dot represents a 10-meter increase. The evidence attests to a sharp improvement of the predictive power of the model as the spatial resolution declines when starting from a $10\times 10$-meter grid. Note that on the $10\times 10$-meter grid, our model explains only about 31\% of the variation in true building heights in the reference data. At the $200\times 200$-meter grid, the explanatory power increases to over 80\%. The R-squared converges to around 98\% after $2000\times 2000$-meter aggregation. This reflects how well the proposed model predicts actual building heights on the spatial meso scale. Given the image quality of Sentinel as discussed in Section \ref{sec:case}, one often focuses on medium-aggregated levels (such as ones on a $200\times 200$-meter grid) of the floorspace density, as the latter offers sufficiently-precise indicators of urban development. We illustrate in Figure \ref{fig:scatter_rf_pd} the correlation of predicted and targeted heights (on a log scale) in the trustworthy reference data when the height distribution is aggregated to $200\times 200$-meter cells. An inspection of this figure attests to the strong correlation suggested by Figure \ref{fig:agg-corr}.}

\subsection{Correlations with Nightlight Data}

\label{sec:nlt-check}

\begin{figure*}[!t]
    \centering
    \includegraphics[width=0.8\linewidth]{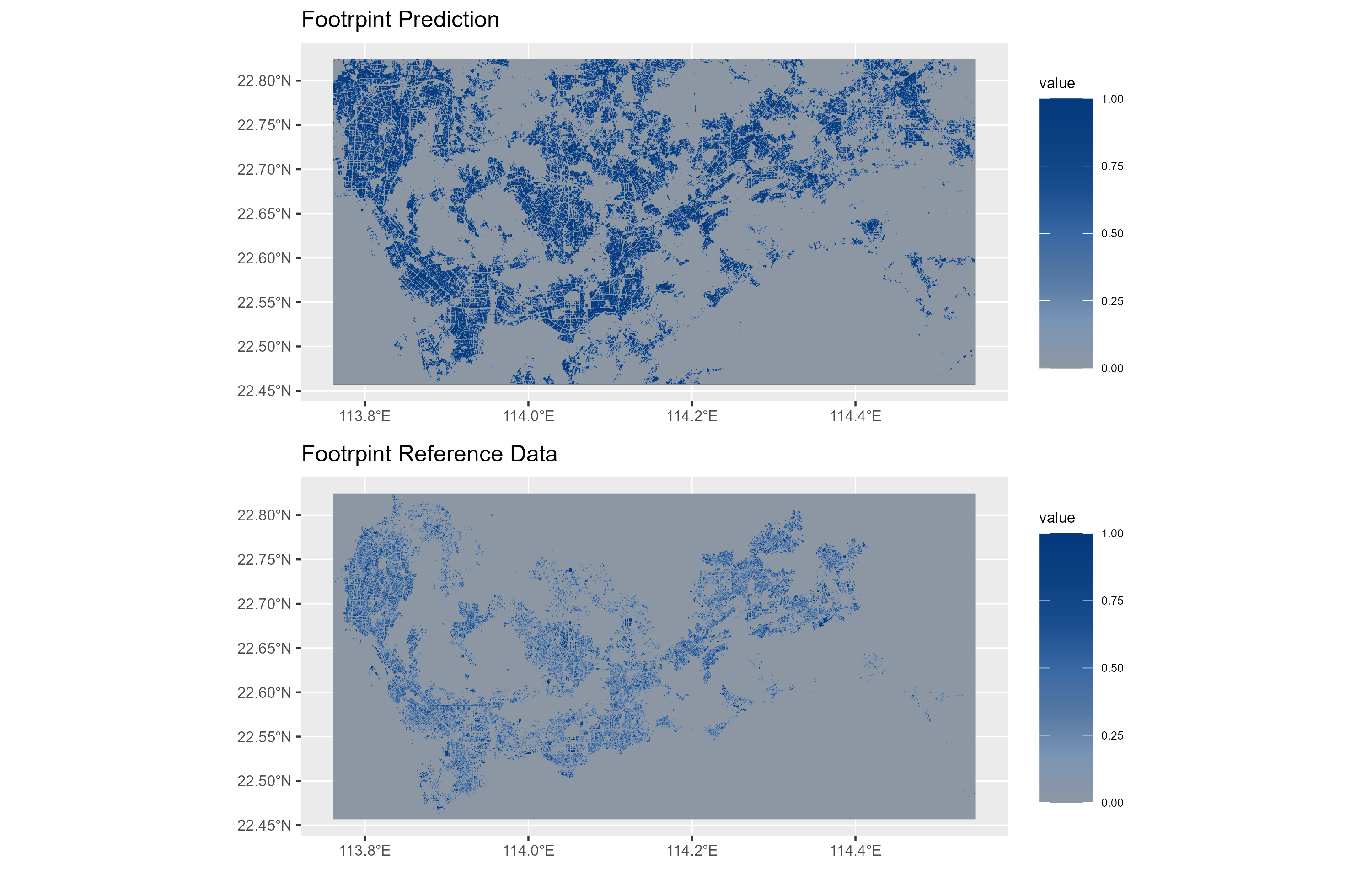}
    \caption{Footprint of All Reference Data and Model Predictions for Shenzhen 2019.}
    \label{footprint_rf_pd_maps}
\end{figure*}

\begin{figure*}[!t]
    \centering
    \includegraphics[width=0.8\linewidth]{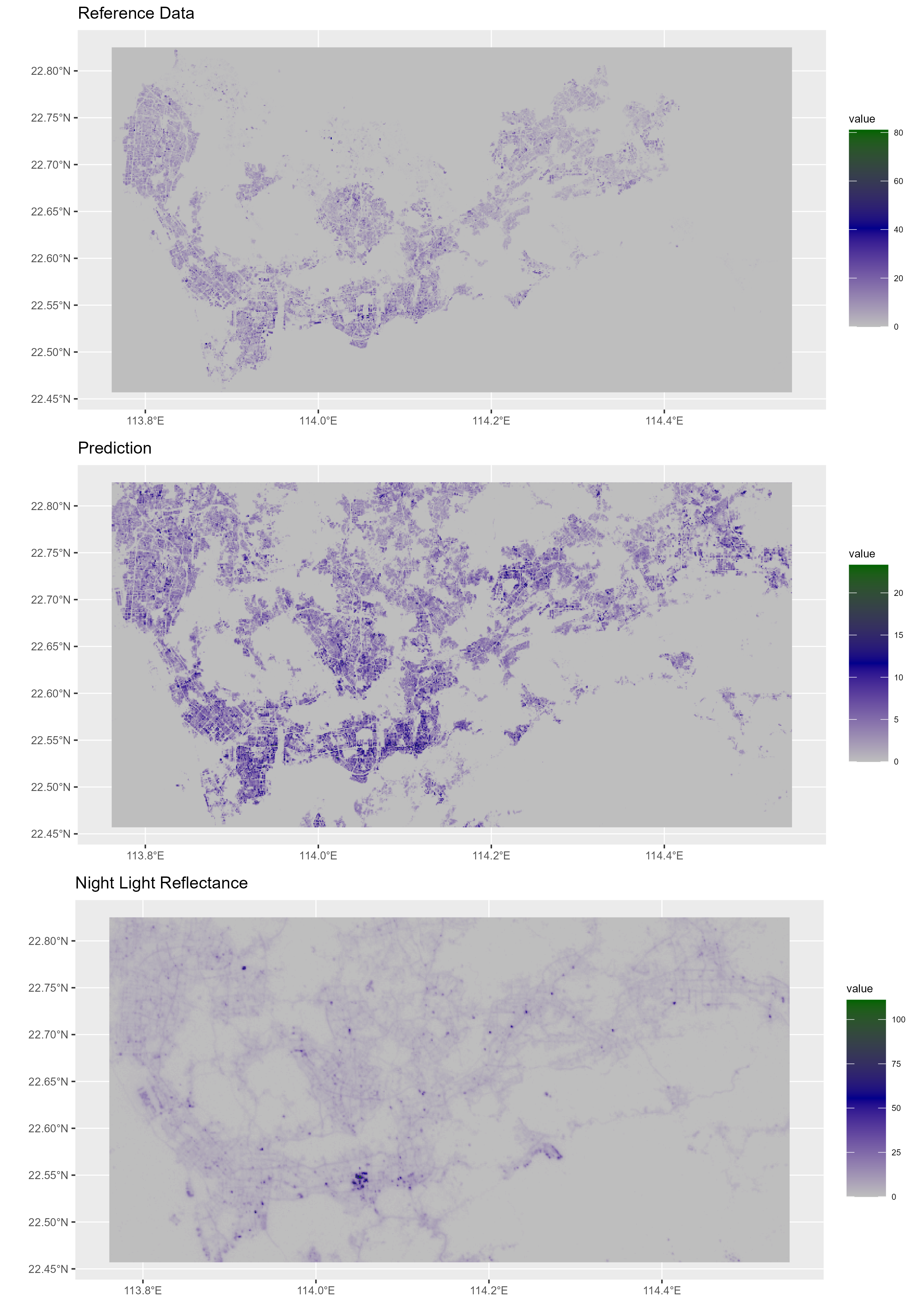}
    \caption{All Reference Data, Model Predictions, and Luojia NTL Data for Shenzhen 2019.}
    \label{rd_pr_nlr_map}
\end{figure*}

\begin{figure*}[!t]
    \centering
    \includegraphics[width=0.8\linewidth]{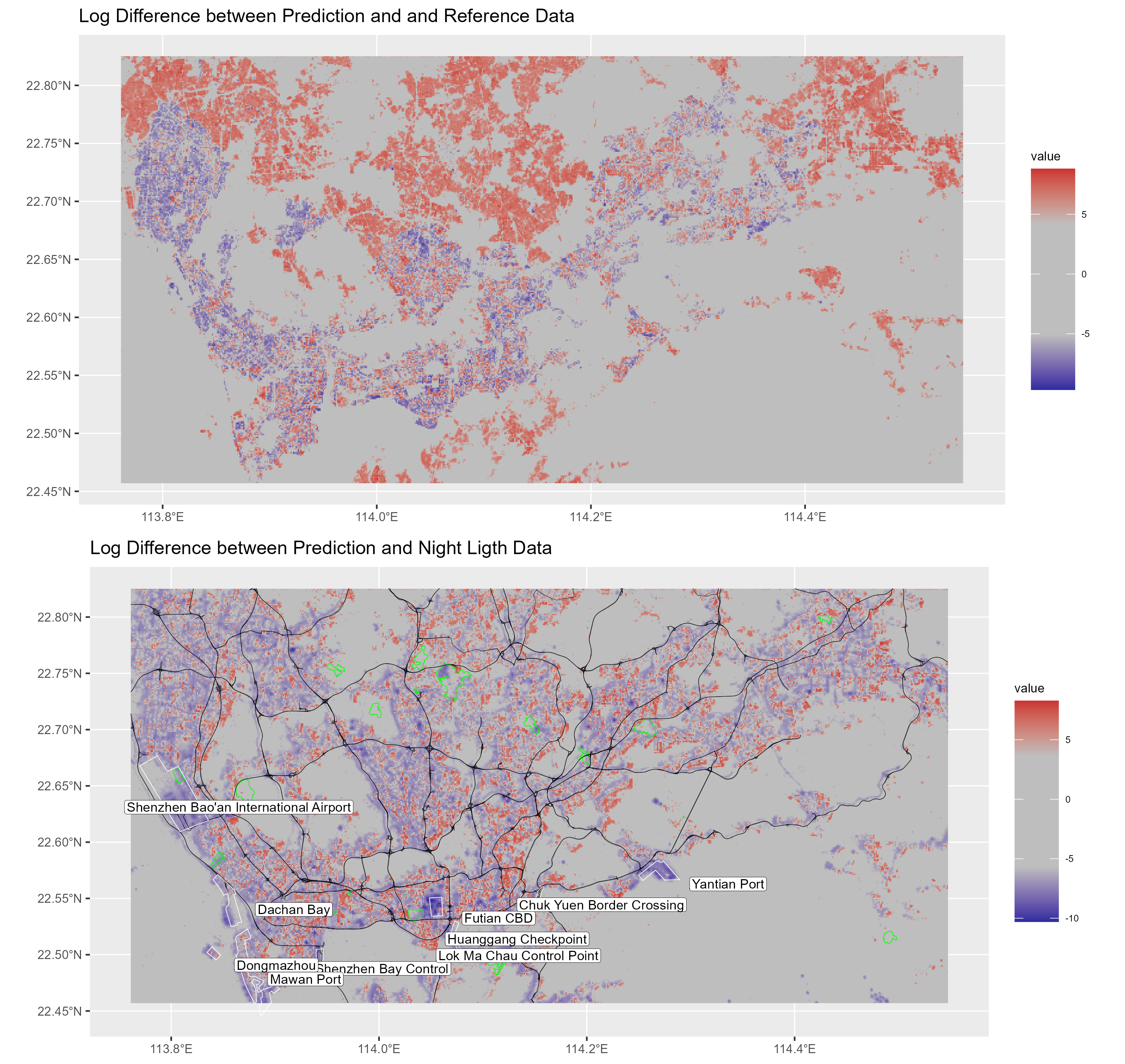}
    \caption{Log-difference Map between Model Predictions and All Reference Data and Model Predictions and Nightlight Data. Red zones indicate an over-prediction of our model compared to the reference data and an under-prediction of the NTL data relative to our model. Purple zones indicate an under-prediction of our model relative to the data and an over-prediction of the NTL data relative to our model. Black lines: highways from OpenStreetMap (OSM), green lines: golf courts, white lines: selection of infrastructure with nightlight radiance.}
    \label{fig:diff}
\end{figure*}

\new{Here, we look at all the reference data points in Shenzhen and the predictions our model for the whole city area. In the footprint prediction shown in Figure \ref{footprint_rf_pd_maps}, our model covers all the city areas that are missing in the current reference data. }

\new{Figure \ref{rd_pr_nlr_map} shows the comparison of three data samples of Shenzhen, all the reference data in Shenzhen, our prediction by the best performer and Luojia 1-01 NTL data \citep{deren2019design} radiance intensity. The reference data and our prediction are shown with the average heights per cells aggregated to a uniform $120 \times 120$-meter scale for consistency with the NTL data. The illustration clearly indicates that the three data sets describe a similar urban structure. A robust correlation exists between the NTL radiance and the predicted regional building volume.}

\new{Figure \ref{fig:diff} shows the difference between our prediction multiplied by the respective correlation coefficient and all the reference data in Shenzhen (upper map) and the NTL data (lower map). Both differences are scaled by their standard deviation and logarithmized. The lower map also shows highways (in black) as found by OpenStreetMap (OSM), a selection of infrastructure features (in white) and golf courses (in green). Red zones indicate overestimation of our prediction compared to the reference data and an under-prediction of the NTL data, while purple zones indicate under-prediction relative to the reference data and an over-prediction of the NTL data. The upper map shows that regions such as Hong Kong in the lower middle or peripheral areas around the edges of the map that are clearly indicated by NTL but are missing from the reference data, are featured in our prediction. The lower map uncovers some shortcomings of the NTL data. The latter heavily over-represent large-scale infrastructures that are illuminated during the night, such as ports, customs checks, airports, and highways as well as recreational facilities such as golf courts. On the contrary, NTL data underestimate building volumes in central business districts (CBD) with more high-rise buildings as well as in other very dense urban environments.}

\new{The deviation between building volumes and the NTL intensity underlines the need of a robust pipeline of the proposed sort to improve research that so far has been relied mainly on NTL data. We see from figure \ref{fig:diff} that our measure and NTL data highlight different aspects of the urban environment. The former gives a better overview of the building and floorspace density, while the latter emphasises large scale infrastructures and roads. Especially, approximations of population density and building structure measures that require a more fine-grained spatial resolution cannot be done without much bias based on NTL data. In conclusion, NTL data, typically derived from satellite imagery, can provide insights into areas with high economic activity but may not capture well residence-based and buildings- as well as floor-space-related measures of urban density.}

\section{Related work}
Here, we summarize relevant earlier work in the domain of remote sensing that has inspired our contribution. 

\textbf{2D-mapping.}
Remote sensing methods of various kinds have been in use for a long time in earth-science disciplines such as ecology, glaciology, and geology since the start of the first Landsat missions 1972. In the 2000s, thanks to increased data quality and improved artificial intelligence methods, renewed interest in remote sensing for urban structures had arisen. It became increasingly possible to map the whole globe in land use categories, although in a relatively low resolution of 500 meter cells \citep{SCHNEIDER20101733}. Related research is interested in mapping different land use types, where urban use is just one among others. The possible resolution was step-by-step refined to up to 13 meters \citep{ESCH201730} and 10 meters \citep{GONG2019370} for global mappings. At this granularity, individual buildings become visible, and one can sensibly speak of building footprints beyond mere land-use classifications. While classification in the former research is based on unsupervised learning methods and a random forest framework in the latter, \citet{WU2021112515} provide building-footprint mappings for a selected set of cities of China based on a U-Net deep learning architecture. 

\textbf{Building-height estimation.}
In the past few years, a new strand of literature has emerged, which estimates building height and the intensity of land use.   
\citet{LI2020111859} create a 3D continental mapping of North America, Europe, and China, using a random-forest model. The main input sources for the estimation are Landsat 8 (optical) and Sentinel-1 (SAR), and the final data set has a resolution of roughly $1\times 1$ kilometers. \citet{FRANTZ2021112128} use a support vector machine method based on Sentinel-1 (SAR) and Sentinel-2 (optical) to obtain building-height estimates based on shadow length and radar reflectance. The result of the latter project is a national map of Germany with a $10\times 10$-meter resolution. \citet{CAO2021112590} build on in terms of resolution vastly superior multi-view image data from the Chinese Ziyuan-3 satellite. Using a deep learning methodology, they produce a 3D mapping with a $2.5\times 2.5$--meter resolution for a set of 42 Chinese cities. Yet, neither the satellite imagery nor the resulting mapping are available on open-source bases. Further recent contributions that have looked at building-height estimates in China include \citet{rs12223833}, \citet{yang_building_2022}, and \citet{yang_building_2022}.
\citet{yuetal2021} generate a panel of building heights for Shenzhen only that runs from 1986 to 2017. They start from a high-resolution Light Detection and Ranging (LiDAR) map, and reversely update with low-resolution Landsat archive imagery. \new{\citet{li2023shafts} use  CNN in a multi-task setting to study building footprint and height of 38 cities worldwide, where they have good 3D reference data. They report an $R^2$ of building-height and -footprint prediction which exceed those of traditional ML models by 0.27–0.63 and 0.11–0.49, respectively.} 

\section{Discussion, future work, and conclusions} \label{sec:future}
\textbf{Discussion: data quality.}
Using data of a meso-grained resolution of 10 meters provides an advantage of large cross-sectional and time-series coverage. However, it comes at an inevitable cost of precision. At this stage, we provide our dataset and pipeline as an open-source platform. However, in the future, we will harmonize various satellite images including Planet, which would substantially increase building visibility. \footnote{Our team is supported by ETH Zurich's library which has acquired an academic licence from Planet Labs PBC. We cannot open-source the Planet images, but we will provide pre-trained models trained by harmonized Planet-plus-Sentinel images.}

\textbf{Discussion: better reference data of building footprint and height.} 
Our team is actively developing a tool that crawls the up-to-date Amap 3D reconstruction.\footnote{See \url{https://lbs.amap.com/demo/javascript-api/example/3d/fixed-view} for an official static demo provided by Amap.} We provide a short illustration of our tool in Appendix~\ref{app:ground-truth}. This set of reference data should be of better quality and can, hence, save image-annotation efforts in the absence of open-source reference data.

\textbf{Discussion: annotation.}
The idea is that for areas of interest without reference data (via Amap), we need to acquire human annotations. 
We are in contact with professionals who are experienced in satellite image annotations. We can benefit from their expertise and tool designed to annotate forest data.\footnote{Our contacts at Restor, see \url{https://restor.eco/about/team/}. }  

\textbf{Discussion: robustness checks.}
We will carry out robustness checks at an aggregated spatial level with demographic indicators of cities in census data, for example, on building coverage across Chinese counties/cities. 

\textbf{\new{Discussion: task coefficients in the multi-task learner.} }
\new{In future work, it is worth exploring the coefficients of various tasks, which can be learned from the data. Our current approach relies on a grid-search of pre-defined coefficients, chosen based on the numerical scales of tasks. \citet{CAO2021112590} proposed using learnable coefficients (task uncertainty) to parameterize each task, following \citet{kendall2018multi}. This approach would enable our network to automatically and dynamically learn the weights assigned to each task without manual tuning efforts.}

\textbf{\new{Discussion: time-series data.}}
\new{One of the merits of the proposed approach will be that it can be used in conjunction with medium-resolution remote-sensing data that permit tracking changes in floorspace for longer time windows. The latter is elemental not only for assessing hypotheses in urban studies (related to the determinants as well as the effects of agglomeration on economic outcome) and public and transport economics (related to the effects of infrastructure developments and agglomeration) but also for statistics and econometrics, if identification of the relationships and of causal effects is sought from changes in the data rather than from cross-sectional states thereof. We plan to work on the cross-sectional combination and time-series concatenation of several sources of satellite imagery to produce consistent data panels that are longer than life-cycles of satellites.}

\textbf{Conclusions.} \new{
This paper provides a first milestone in generating a rich dataset on building-specific floorspace (that is, building footprint and height) and a pipeline to train and empirically evaluate the floorspaces. We use a single multi-task object learner to collectively learn building footprints and heights. We have also analyzed the quality of prediction in aggregated urban units, as well as its correlation with other urban development data like nightlight. Our paper lays the groundwork for further empirical research that builds on the theoretical foundation of quantitative urban economic models. In the future, we plan to extend the pipeline to all Chinese cities across the time span of 2009-2022.}

\begin{acks}
We gratefully acknowledge the extensive help and excellent support of our research assistants Muyang Jiang, Sapar Charyyev, Yuru Jia, and Dal Fergus throughout the project. Special thanks go to our collaborator Dr.~Josh Veitch Michaelis at Restor/DS3Lab, for his inspirational discussions and brainstorming sessions.  
\end{acks}
\bibliographystyle{ACM-Reference-Format}
\bibliography{sample-base}
\clearpage
\appendix

\section{Model architecture} 
\label{app:resnet-34}

ResNet-34 performs best in the U-Net backbone experiments described in Section \ref{sec:eval} and its illustrated architecture is as follows. 
\begin{figure*}[!h]
    \centering
    \includegraphics[width=1\linewidth]{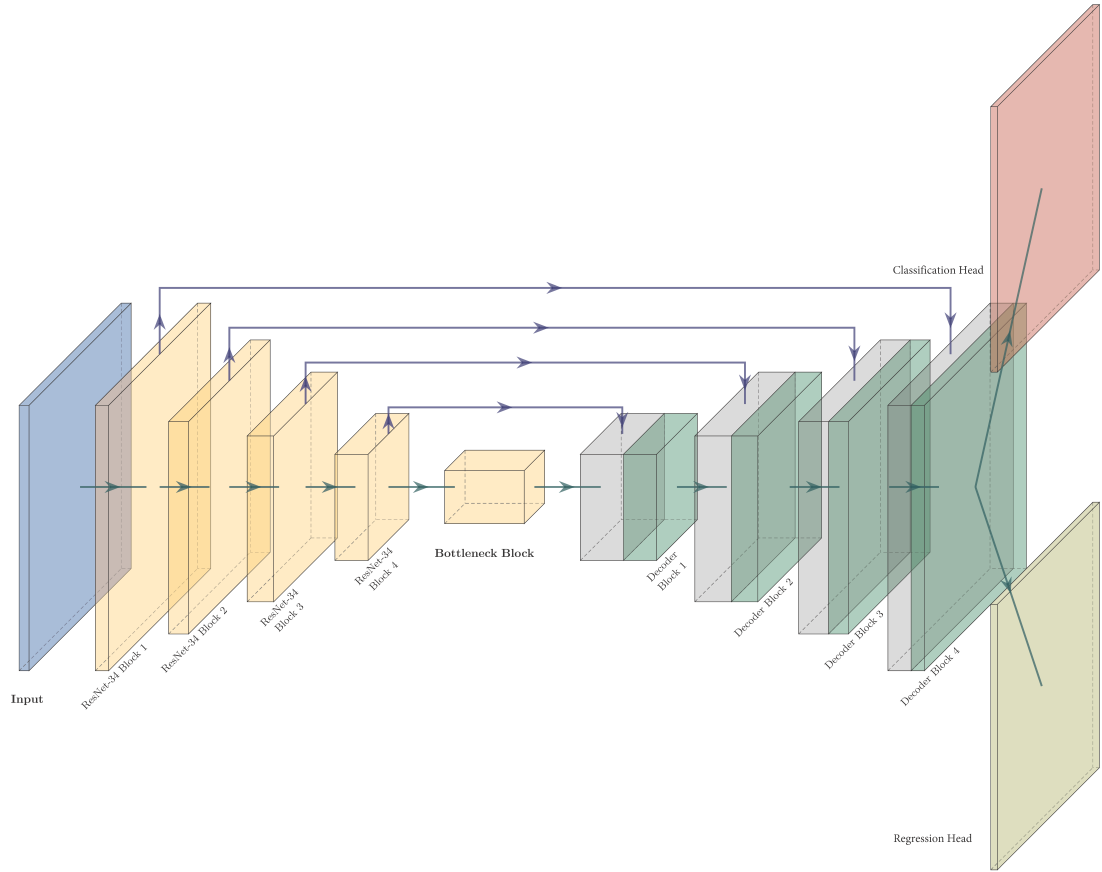}
    \caption{Model Architecture with ResNet-34 Backbone.}
    \label{fig:model_arch}
\end{figure*}

\section{More results on building footprint prediction}
\label{app:footprint}
In our previous study \cite0{egger2023building}, we have carried out experiments with different setups of U-Net to predict footprint as shown in Table~\ref{tab:metrics-old}: (a) Sentinel-1 (SAR data), (b) Sentinel-2 (optical images), and (c) Sentinel 1+2 (the overlay of Sentinel-1 and -2). We see that combining SAR data with optical images in (c) gives us the best performance in all metrics. We have four classes in the \textit{softmax} function, that is, no building, buildings with a height of below 6 meters, 6 to 27 meters and above 30 meters.\footnote{We have tested different cutoff values, e.g., below 10 meters, 10 to 20 meters, above 20 meters, the results are very close, though. } These cutoffs are chosen to slice the multi-modal distribution of building heights in all 77 cities in the crowd-sourced dataset into three categories of a local maximum (52\%, 39\%, 8\%). \new{Note that in our multi-task learner, we did not use the cutoff values but we build a regression-based model to predict the heights.} 
\begin{table*}[!h]
\centering
\resizebox{0.71\textwidth}{!}{
\begin{tabular}{cccccccccc}
\toprule
                   & \multicolumn{3}{c}{\textbf{(a) Sentinel-1}} & \multicolumn{3}{c}{\textbf{(b) Sentinel-2}} & \multicolumn{3}{c}{\textbf{(c) Sentinel 1+2}} \\
                   \cmidrule(lr){2-4} \cmidrule(lr){5-7}
                   \cmidrule(lr){8-10}
                   & Accuracy   & Precision   & Recall   & Accuracy   & Precision   & Recall   & Accuracy    & Precision    & Recall   \\
                   \cmidrule(lr){2-4} \cmidrule(lr){5-7}
                   \cmidrule(lr){8-10}
\textbf{Footprint}          &       0.722&            0.320&          0.598&       0.741&             0.336&          0.560&        \textbf{0.762}&\textbf{0.375}              &\textbf{0.617}         \\

\cmidrule(lr){1-4} \cmidrule(lr){5-7}
                   \cmidrule(lr){8-10}
\textbf{Height} &       0.670&            0.194&           0.364&       0.703&             0.229&          0.382&        \textbf{0.724}&\textbf{0.261}              &\textbf{0.449}     \\
\bottomrule
\end{tabular}
}
\caption{U-Net Performance Metrics with Sentinel-1 and -2.}
\label{tab:metrics-old}
\end{table*}

\section{Robustness check with two-stage models}
\label{app:2-stg}
\textcolor{black}{To validate the robustness of our proposed one-stage model described in \citep{egger2023building}, we have tested two-stage models with Sentinel 1 + 2 images. In the two-stage models, we first predict only the footprints of the buildings (the first stage) and then feed the predicted footprints into a height prediction model (the second stage). We have validated three architectures in the second stage of the two-stage models when predicting heights: (A1) stacking the building footprint prediction results and the Sentinel images together as height prediction model inputs; (A2) clipping Sentinel images with building footprint prediction results before putting them into the height prediction model; (A3) training in parallel the building footprint prediction and height prediction in the encoder part, and use skip connection to concatenate them for the bottleneck and decoder parts of the U-Net model as in \citet{CAO2021112590}. The first stage (footprint prediction) is identical among these three architectures. The loss and activation functions of each stage are similar to those used in the one-stage model; \textit{sigmoid} and \textit{softmax} as activation for footprint and height predictions, respectively; the loss function remains the weighted Dice coefficient. } 

\textcolor{black}{The results of the two-stage models show a slight increase in the prediction of building footprints, as we see in Table~\ref{tab:robustness}, but the height prediction performance remains almost the same for all of our attempts. The first stage (footprint) prediction takes 10 seconds on average per epoch, while A1, A2 and A3 take 17.5, 20, 18.75 seconds, respectively, in the second stage (height prediction). Compared to our segmenter described in \citep{egger2023building} (with 11 seconds per epoch), which learns the footprint and height in a one-stage model, the computational cost of two-stage models is substantially higher.} 


\begin{table}[!h]
\centering
\resizebox{0.7\linewidth}{!}{
\begin{tabular}{cccccccc}
\toprule
                   & \multicolumn{7}{c}{\textbf{(c) Sentinel 1+2}} \\
                   \cmidrule{2-8}
                   & \multicolumn{3}{c}{\textbf{Footprint}} & \multicolumn{4}{c}{\textbf{Height}} \\
                   \cmidrule(lr){2-4} \cmidrule(lr){5-8}
                   & Accuracy    & Precision    & Recall & Accuracy    & Precision    & Recall & F1\_micro  \\
                   \cmidrule(lr){2-4} \cmidrule(lr){5-8}
\textbf{One-stage}          & {0.762}&{0.375}              &{0.617}         & 0.724&0.261              &\textbf{0.449}  & 0.330   \\
\cmidrule(lr){1-4} \cmidrule(lr){5-8}
                   
\textbf{Two-stage (A1)} &         \multirow{3}{*}{\textbf{0.773}}     &         \multirow{3}{*}{\textbf{0.442}}              & \multirow{3}{*}{\textbf{0.673}}  & \textbf{0.726}   & 0.261  & 0.433  & 0.326  \\
\textbf{Two-stage (A2)}                  &   &  &  & 0.725  & 0.253 & 0.428 & 0.318    \\
\textbf{Two-stage (A3)}                    &   &  &  &  0.725  & \textbf{0.266} & 0.446  & \textbf{0.333}   \\
\bottomrule
\end{tabular}
}
\caption{Robustness Check of U-Net Performance Metrics with Sentinel 1+2 in Two-Stage Models.}
\label{tab:robustness}
\end{table}

\section{Weighted log loss and its results}\label{app:weighted_loss}
\begin{figure*}[!h]
    \centering
    \includegraphics[width=0.8\linewidth]{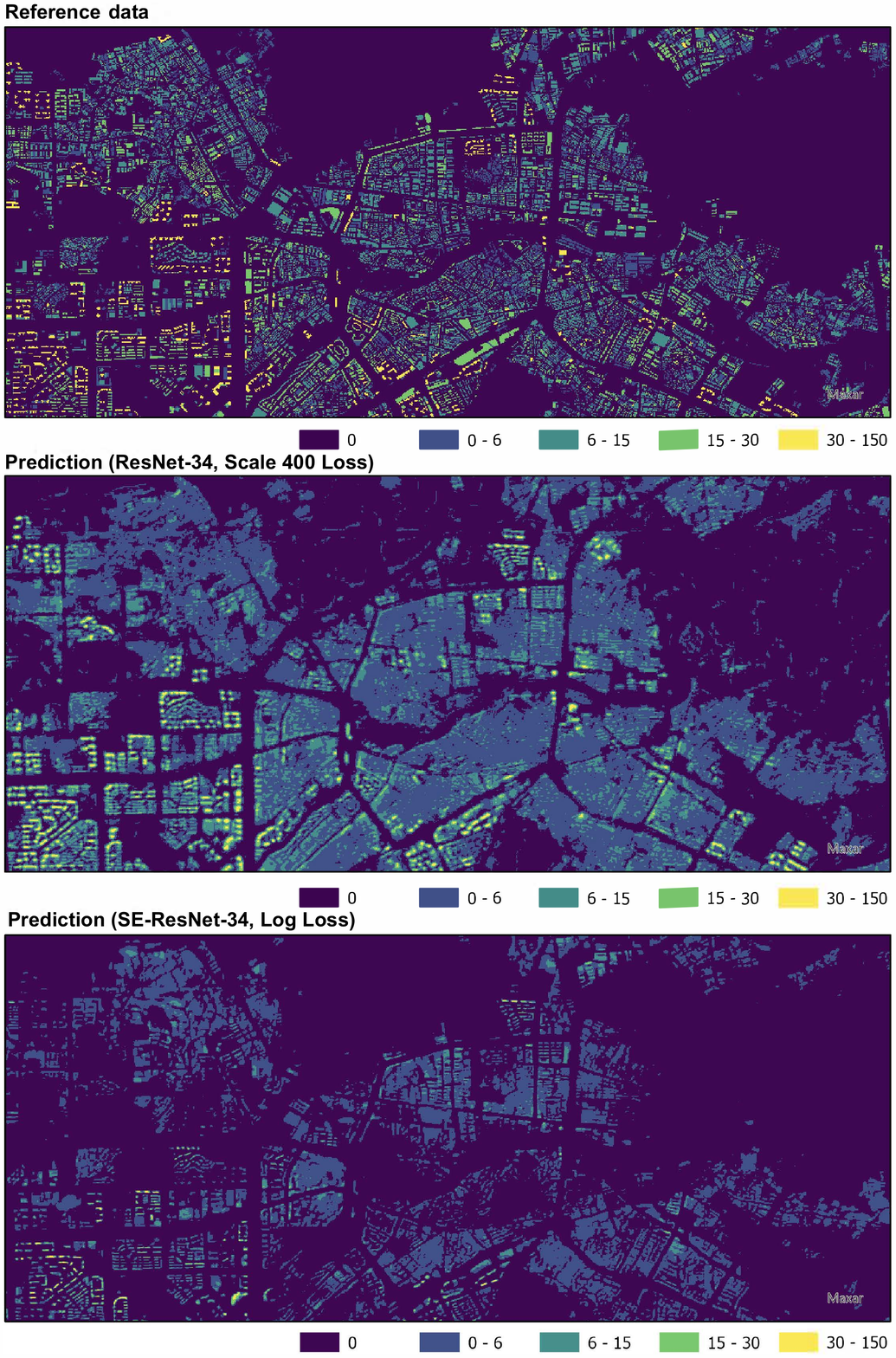}
    \caption{Visualization of Two Models' Prediction in Shenzhen.}
    \label{fig:comp-pred}
\end{figure*}

\begin{figure*}[!h]
    \centering
    \includegraphics[width=1\linewidth]{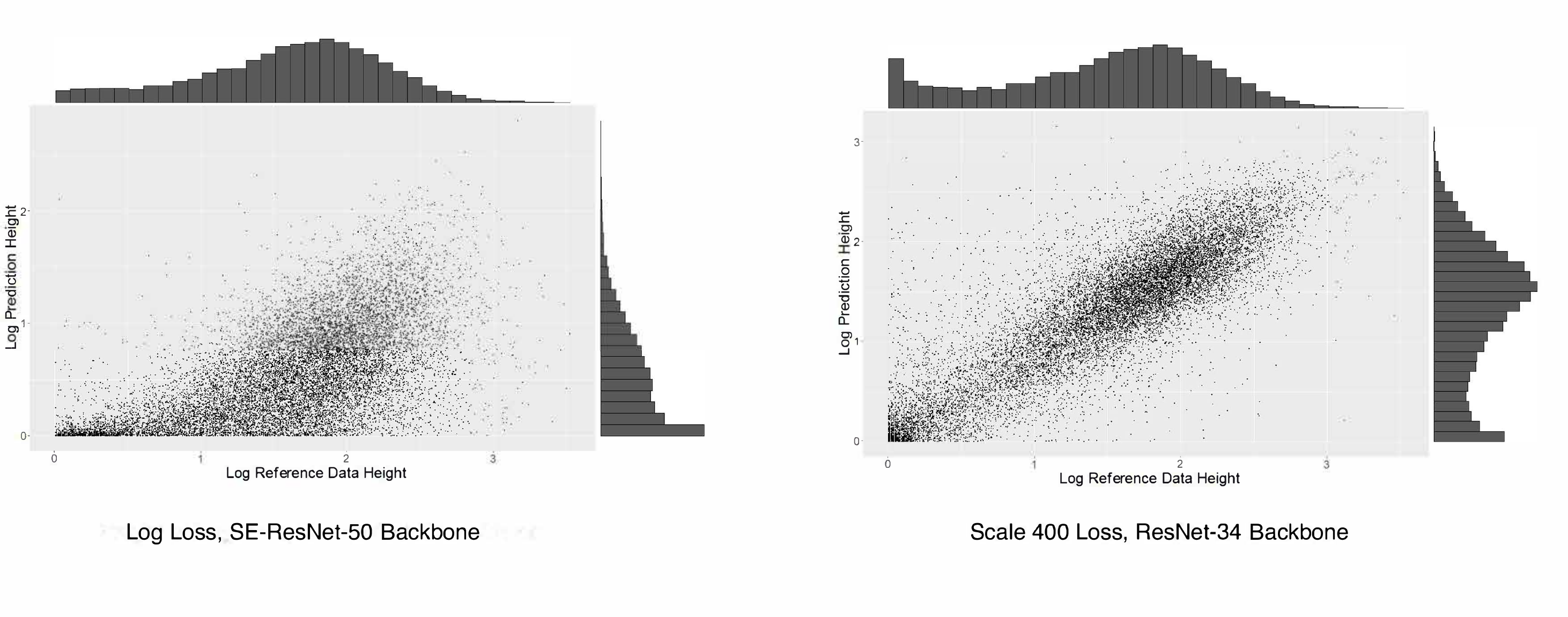}
    \caption{Comparing Scatter Plots of Aggregated to 200m Cells Logarithmic Mean Height Plus One.}
    \label{fig:hist_comp}
\end{figure*}

\begin{figure*}[!h]
    \centering
    \includegraphics[width=1\linewidth]{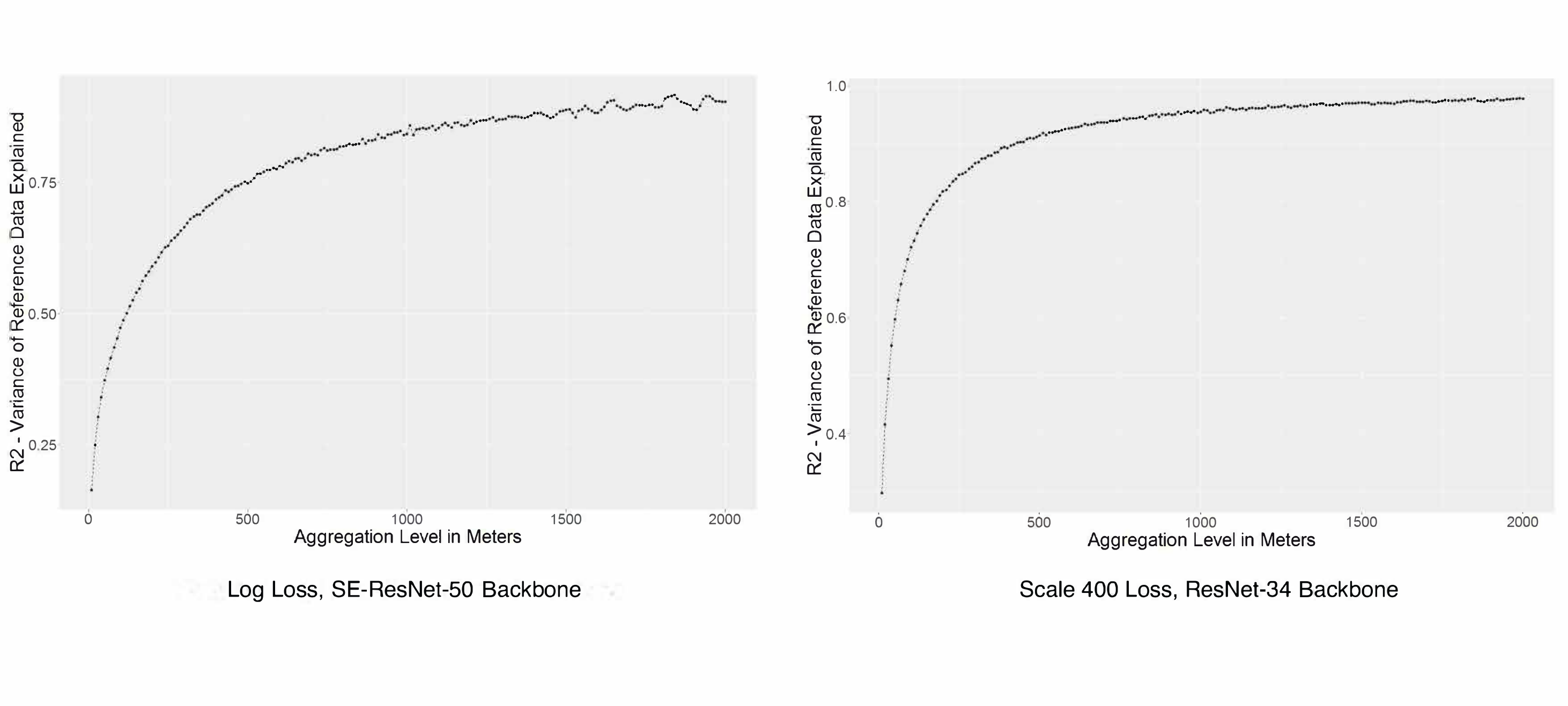}
    \caption{Comparing Share of Explained Variance of Building Height in the Reference Data by Model Predicted Building Height as a Function of the Aggregation Level.}
    \label{fig:agg_comp}
\end{figure*}

\new{We compare the two setups, ResNet-34 (scale 400 loss) vs.~SE-ResNet-50 (log scale loss) in predicting footprints and heights in Shenzhen. We look at their predictions in Shenzhen on various height classes and footprint coverage (Figure \ref{fig:comp-pred}), the correlation of predicted and target heights when aggregating to 200m cells (Figure \ref{fig:hist_comp}, as well as the share of explained variance of building height in the reference data by model-predicted building height as a function of the aggregation level (Figure \ref{fig:agg_comp}).}

\section{Better reference data with Amap}
\label{app:ground-truth}

\begin{figure*}[!h]
    \centering
    \includegraphics[width=\linewidth]{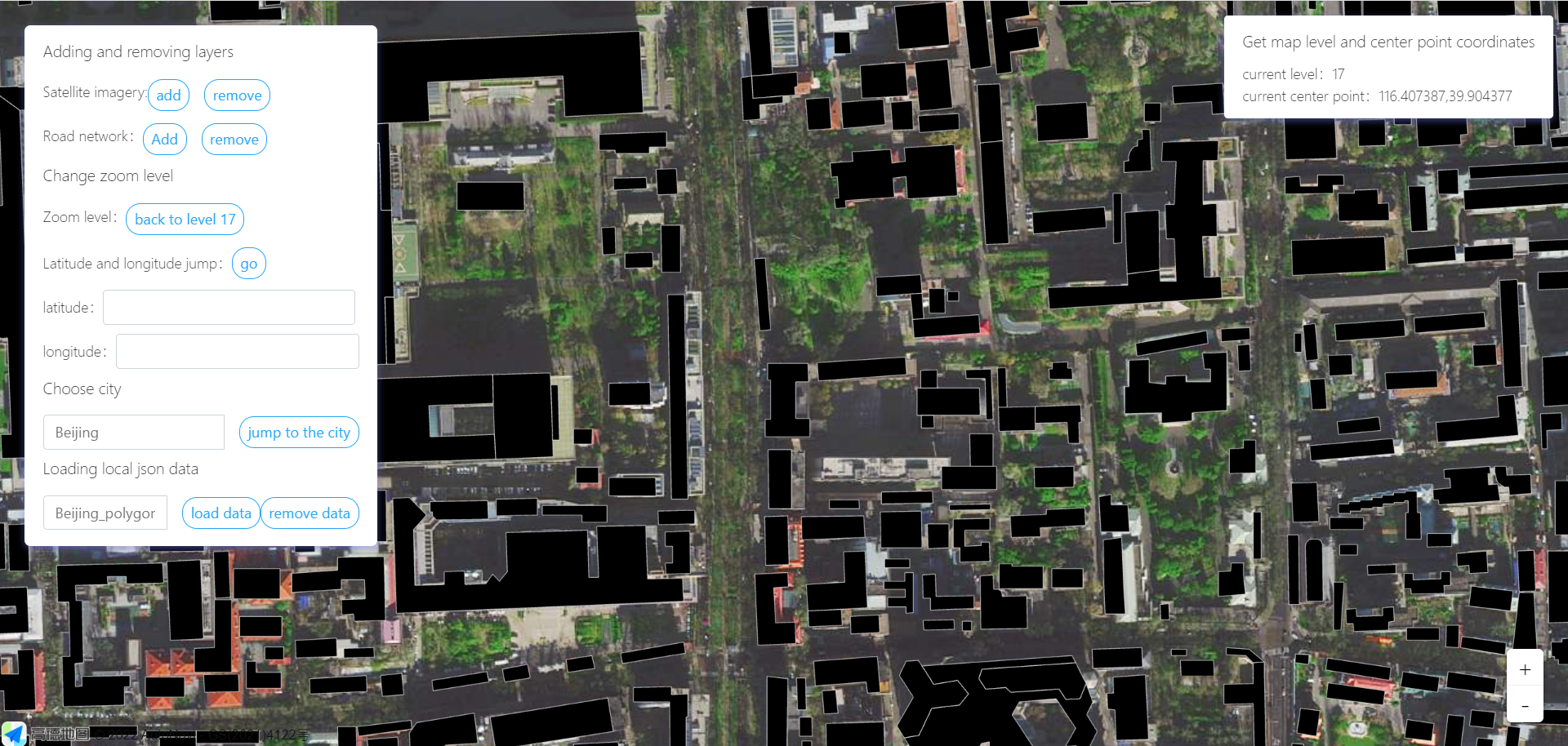}
    \caption{Customized Toolbox for Accessing Amap's Building Footprint and Satellite Imagery Dataset, Built on Amap's API.}
    \label{fig:tool}
\end{figure*}

\begin{figure*}[!h]
    \centering
    \includegraphics[width=\linewidth]{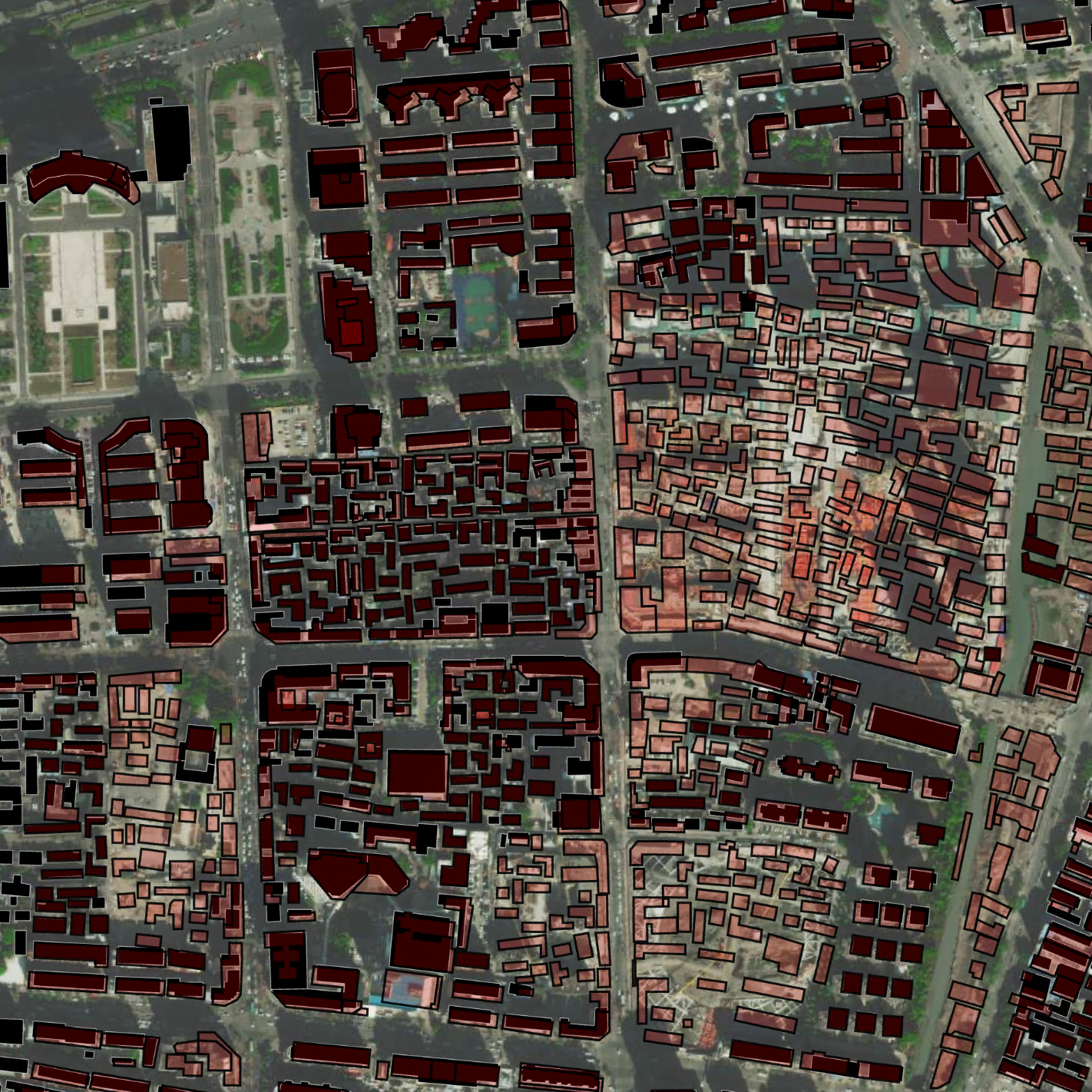}
    \caption{Comparison of the Building Footprint in Crowd-Sourced Dataset (Red) and Up-to-date Amap Dataset (Black).}
    \label{fig:example}
\end{figure*}

Figure~\ref{fig:tool} is a screenshot of our customizable tool built on top of Amap's API. By querying a polygon json, we obtain the actual footprint ground truth provided by Amap. In Figure~\ref{fig:example}, we contrast the ground truth provided by the actual Amap (in black) and the reference data in the crowd-sourced dataset (in red, supposedly from about 2017). These two sets of reference data are overlaid on top of a satellite image of the latest high-resolution image of a sampled area. The two satellite images used here are provided by Amap in the period 2021-2022. According to Amap, the source is DigitalGlobe\&spaceview, with an update frequency 0.5-1 year for urban areas, longer for remote areas. As we can see in the mismatched area, the actual ground truth data better represent the building coverage. In future work, we will systematically evaluate the quality of various reference data. Moreover, we also see some areas of interest where no reference data is available; therefore, minimal human efforts to annotation are expected to bring benefits as we discussed in Section~\ref{sec:future}.

\end{document}